\documentclass[11pt,a4paper]{article}
\usepackage[hyperref]{ranlp2023}
\usepackage{booktabs}
\usepackage{times}
\usepackage{latexsym}
\usepackage{geometry}%
\usepackage{multicol}
\usepackage{graphicx}
\usepackage{calligra}
\usepackage[IL2]{fontenc}

\usepackage{amsmath}

\usepackage{microtype}

\aclfinalcopy 

\title{Cordyceps@LT-EDI: Patching Language-Specific Homophobia/Transphobia Classifiers with a Multilingual Understanding}

\author{Dean Ninalga \\
\texttt{justin.ninalga@mail.utoronto.ca} 
}
\date{}

\begin{document}
\maketitle
\begin{abstract}

Detecting transphobia, homophobia, and various other forms of hate speech is difficult. 
Signals can vary depending on factors such as language, culture, geographical region, and the particular online platform.
Here, we present a joint multilingual (M-L) and language-specific (L-S) approach to homophobia and transphobic hate speech detection (HSD). 
M-L models are needed to catch words, phrases, and concepts that are less common or missing in a particular language and subsequently overlooked by L-S models.
Nonetheless, L-S models are better situated to understand the cultural and linguistic context of the users who typically write in a particular language. 
Here we construct a simple and successful way to merge the M-L and L-S approaches through simple weight interpolation in such a way that is interpretable and data-driven.
We demonstrate our system on task A of the \emph{Shared Task on Homophobia/Transphobia Detection in social media comments} dataset for homophobia and transphobic HSD.
Our system achieves the best results in three of five languages and achieves a 0.997 macro average F1-score on Malayalam texts.
\end{abstract}

\section{Introduction}
In general, the US is seeing an increase in institutionalized transphobia in the form of banning gender-affirming care and the banning of transgender youth from several sports \cite{Kline2023MappingTP}. 
However, studies on individuals who experience institutionalized transphobia in the US experience more psychological distress and instances of suicidal ideation \cite{Price2023StructuralTI}. 
The codifying of anti-trans laws then certainly must give confidence to those with transphobic beliefs and desires to spread anti-trans rhetoric in online spaces. 
\citet{Berger2022SocialMU} recently presented results showing that LGBTQ youth often rely on social media for improved mental health outcomes and as a source of social connection that helps close mental health disparities. 
Therefore, appropriate content moderation on social media platforms stands to benefit from accurate NLP systems that can identify homophobia, transphobia, and other forms of hate speech.

Good knowledge of hate speech in a particular language may not always be useful for other languages, yet many common phrases and sayings are often expressed across languages.
Namely, purveyors of hate speech often do not openly say hateful comments but instead rely on equally vicious code phrases, or \emph{dogwhistles}, to avoid existing content moderation systems \cite{Henderson2017HowDW, Magu2017DetectingTH}.
Knowledge of the hidden meanings of these encoded sayings can create powerful tools for improving online moderation \cite{Mendelsohn2023FromDT}.
These phrases can easily transcend the regions of their origin, spreading across online communities without detection in vulnerable communities.
Hence, knowledge of dogwhistles in their current form will make content moderation systems more robust to these signals as they appear in different languages in new online spaces.

Textual databases built for hate speech analysis are predominantly in English, which creates language-based performance disparities \cite{Jahan2021ASR, Poletto2020ResourcesAB, Aluru2020DeepLM}.
As \citet{Wang2020OnNI} suggested, in M-L models languages are competing for model resources, potentially resulting in worse performance for low-resources languages.
This performance bias is possibly due to that many M-L datasets used for pretraining popular language models often are majority English samples, often by a wide margin \citep{Barbieri2021XLMTML, Xue2020mT5AM, Ri2021mLUKETP}).
Consequently, there is a general disparity in performance when comparing English-only and M-L HSD models \cite{Rttger2022MultilingualHF}.

\citet{Nozza2020WhatT} push for more pre-trained models in non-English languages as they will (naturally) be best for downstream tasks in the same language domain they are trained in.
However, pre-training techniques typically require large datasets to guarantee good downstream performance.
Given a relative lack of language-specific data for HSP, more indirect and creative approaches are required to alleviate the performance gap between English and non-English tasks.

For our present purposes, we are presented with multiple target languages and tasked to detect levels of homophobia and transphobia for each specified language using an automated system. We introduce Language-PAINT to jointly model M-L and L-S knowledge that incorporates recent work on weight interpolation.

In summary, our main contributions are the following:
\begin{itemize}
\item We publicize a language-based weight interpolation approach as the next step in advancing HSD research.

\item We provide a demonstration of our framework on task A of the \emph{Shared Task on Homophobia/Transphobia Detection in social media comments} \cite{chakravarthi2022can}.

\item We provide preliminary evidence suggesting that our framework is robust to label distribution shifts.

\end{itemize}

\section{Related Work}
\subsection{Language Transfer in Hate Speech Detection}
Several techniques from recent years have worked on closing the performance disparity between majority and minority languages in HSD.
Namely, several attempts directly translate low-resource languages into high-resource ones \cite{Pamungkas2019CrossdomainAC, Ibrohim2019TranslatedVN}.
\citet{Pelicon2021InvestigatingCT} presents a data-based approach that first trains a M-L model for HSD, similar to our training scheme’s initial step. \citet{Pelicon2021InvestigatingCT} use a percentage of L-S  data to finetune their model where the percentage is chosen empirically. 
\citet{Choudhury2017CurriculumDF} delay training with code-mixed data, opting to first train with mono-lingual samples using the two languages used in the code-mixed data.
The popular IndicNLP \cite{Kunchukuttan2020AI4BharatIndicNLPCM} uses bilingual word embeddings for translation and transliteration, typically between English and a target low-resource language.
\citet{Biradar2021HateON} subsequently attempt to incorporate IndicNLP’s \cite{Kunchukuttan2020AI4BharatIndicNLPCM} embeddings for code-mixed HSD.

\begin{figure*}[h]
  \includegraphics[width=\textwidth]{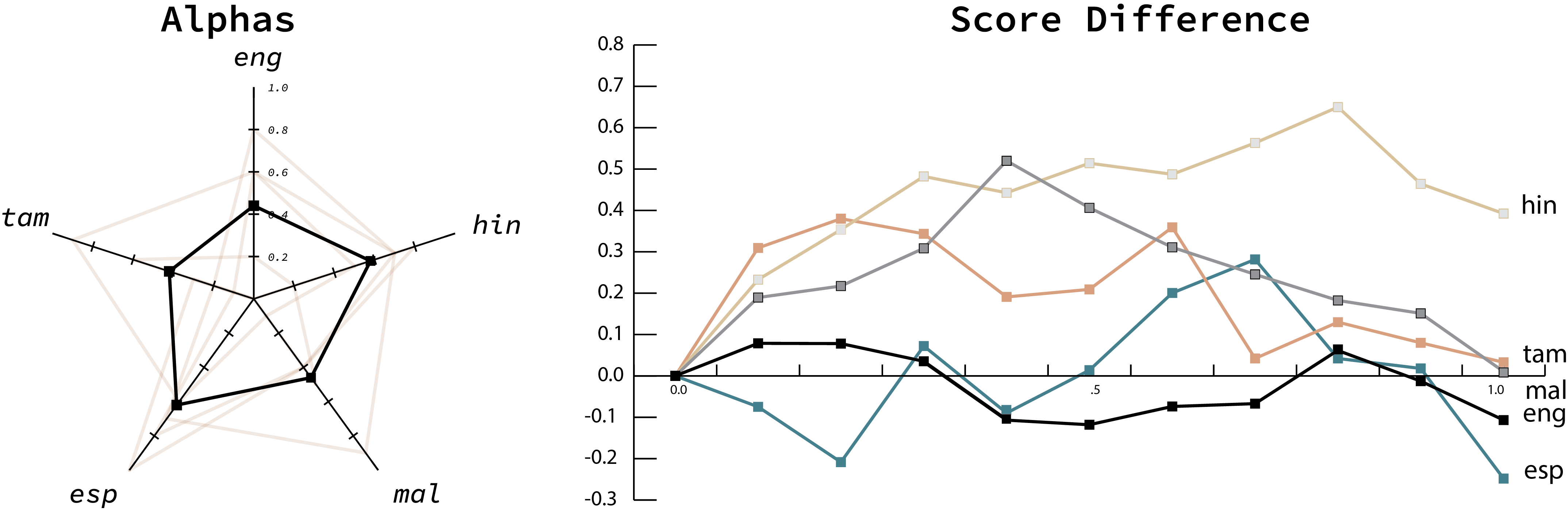}
  \centering
  \caption{\label{fig:alph}
  Left: Average selected value for $\alpha$ (thick black line) averaged over five runs for each language. Right: Average validation F1 score as a function of $\alpha$ reported for each language, averaged over five runs.}
\end{figure*}

\subsection{Weight Interpolation}
In this paper, we adopt the interpolation strategy of \emph{Weight-space ensembles for fine-tuning} (WiSE-FT) \cite{Wortsman2021RobustFO}. In particular, we base our framework on a subsequent variation called PAINT \cite{Ilharco2022PatchingOM} constructed to incorporate the 
input robustness of a zero-shot model into finetuned models across diverse tasks.
Formally, given a single task $t$ takes the weights of the \emph{zero-shot} model $\theta_{z}$ and a finetuned model $\theta_{f}$, the weight interpolation of PAINT performs the interpolation:
$$\theta^t=\alpha\theta_{z}+(1-\alpha)\theta_{f}$$
with $\alpha\in[0, 1]$.
In addition to the specific experiments performed by \cite{Ilharco2022PatchingOM}, recent work shows that averaging two (or more) language models has the potential to leverage knowledge contained in each \cite{Gueta2023KnowledgeIA, DonYehiya2022ColDFC, Choshen2022FusingFM}.
However, no prior work has studied weight space ensembling based on language to the best of our knowledge.

\section{Methodology}\label{methodology}
Here, we use Bernice \cite{delucia-etal-2022-bernice}, a language model exclusively on Twitter\footnote{https://twitter.com} data and is known to be performant on HSD across multiple languages.
Indeed, many studies rely on Twitter, to construct datasets of code-mixed samples for various HSD approaches \cite{Bhat2018UniversalDP, Bansal2020CodeSwitchingPC, Farooqi2021LeveragingTF, Choudhury2017CurriculumDF}, which in aggregate, motivates our choice of language model. 

\subsection{Language-PAINT}
Given $k$ distinct groups of (possibly code-mixed) languages, we first train a M-L model on a dataset that includes all the languages.
We continue training until saturation on a validation set, where we take the average F1 score across languages.
Next, we create an additional $k$ L-S models, - one for each language - where each is initialized with the weights of the M-L model.
Finally, we perform linear interpolation between the weights of the M-L and each of the $k$ L-S models.
The resulting $k$ models are used for inference on each language.

In mathematical terms, Language-PAINT takes the weights of the trained L-S model $\theta^{i}_{ls}$ and the weights of the M-L model $\theta_{ml}$ and performs the following interpolation:
$$\theta^{i}=\alpha\theta^{i}_{ls}+(1-\alpha)\theta_{ml}.$$
Where $\theta^{i}$ is used to create predictions for the respective language $i=1,..,k$ in the test set.
In practice, we select alpha from a discrete set $\alpha\in\{0, 0.1, 0.2,..., 1\}$ and select based on the resulting model’s F1 performance on a held-out validation set.

\subsection{Ensembling}\label{sec:ens}
Our final prediction on the test sets is an ensembled output of five models trained on five stratified folds.
To create these folds, we first conjoined the original training and development sets.
Next, we divided the conjoined dataset into five folds using 80-20 train-validation splits, ensuring we maintain the label distribution across each fold. We then trained a fresh model on each training and validation fold using the methodology that is described above.
For final inference, we sum the output probabilities of the five models selecting the maximum probability as the final prediction.

\subsection{Data Cleaning}
To preserve as much textual information as possible, we apply minimal additional cleaning steps.
Namely, we only remove a sample if it is found to be overlapping in both the train and development data.
In total, we removed 1695 duplicate samples, where 54\% of the dropped samples are in Tamil and 41\% are in Malayalam.

\begin{table}
\centering
\begin{tabular}{c||c}
\hline
\textbf{Paramter} & \textbf{Value}\\
\hline
Batch Size & 16 \\
Learning Rate & 1e-5 \\
Optimizer & Adam \\
Loss & cross-entropy \\
\hline
\end{tabular}
\caption{\label{tab:hyp}
Training Hyper-parameters}
\end{table}

\section{Experiments and Results}
\subsection{Experimental Setup}
Here, we will perform experiments comparing the L-S, M-L, and, LangPAINT approaches. 
For our first experiment, we combine the training and development set into a single case study. We train five models re-sampling a random 80-20 train-validation split for each run and report the average results on the test set.
For our second experiment, we combine the training, development, and, test sets into a single dataset. Where we train ten models re-sampling a random 80-10-10 train-validation-test split for each run, reporting the average of the results on each test set.
For each of our two experiments, we use the \emph{weighted} F1 score to evaluate performance.
All experiments were run on a single Tesla V4 GPU and we provide the training hyperparameters in Table \ref{tab:hyp}.

\section{Results}
The results of our experiments are given in Table \ref{Tab:ts}.
We can see for most languages, the L-S approach tends to perform best, with the exception of the Malayalam language.
This is reflective of our final leaderboard results where we used an ensemble method (see Section \ref{sec:ens}) that achieves a 0.997 macro average F1-score on Malayalam texts.
Additionally, we report the selected values for $\alpha$ and validation score as a function of $\alpha$ in Figure \ref{fig:alph} for this first experiment. 

For our second experiment, our results (see Table \ref{Tab:ts}) are much more in favor of our method.   
Perhaps the considerably worse performance of the L-S and M-S models is due to the high label-distribution shift between the re-sampled train and test splits.
Nonetheless, LangPAINT appears to be robust to this shift and is still able to maintain good performance, with the only exception being the Spanish language.

\begin{table*}
\centering
\begin{tabular}{c|ccc|ccc}
\toprule
 & \multicolumn{3}{c}{\textbf{Test set}} & \multicolumn{3}{c}{\textbf{10 Fold}} \\
\cmidrule(r){2-4} \cmidrule(l){5-7}

\textbf{Language} & L-S & M-L & LangPAINT (ours) & L-S & M-L & LangPAINT (ours)\\
\midrule
eng & \textbf{0.93} & 0.928 & \textbf{0.93} & 0.565 & 0.584 & \textbf{0.94}\\
hin & \textbf{0.943} & 0.939 & 0.939 & 0.478 & 0.541 & \textbf{0.932}\\
mal & 0.965 & 0.97 & \textbf{0.971} & 0.834 & 0.827 & \textbf{0.930} \\
esp & \textbf{0.878} & 0.874 & 0.877 & 0.91 & \textbf{0.932} & 0.877 \\
tam & \textbf{0.927} & 0.923 & \textbf{0.927} & 0.87 & 0.878 & \textbf{0.895} \\
\bottomrule
\end{tabular}
\caption{\label{Tab:ts}
Results of our experiments comparing the language-specific (L-S), multi-lingual (M-L) and, LangPAINT approaches across languages. We report the \emph{weighted} F1 score for each, where the results are the average of five runs.
}
\end{table*}

\section{Conclusion}
In this paper, we introduce LangPAINT. 
LangPAINT is a weight space ensembling strategy \cite{Wortsman2021RobustFO} repurposed to jointly model the multi-lingual and language-specific signals of homophobia and transphobia.
Our experiments suggest that our method is competitive with the language expert models and has the potential to be very robust to label distribution shifts.
On task A of the \emph{Shared Task on Homophobia/Transphobia Detection in social media comments} \cite{chakravarthi2022can} achieving the best results in three of five languages and
achieves a 0.997 macro average F1-score
on Malayalam, a low-resource language.

\bibliographystyle{acl_natbib}
\bibliography{ranlp2023.bib}


\end{document}